# Deep learning-based ecological analysis of camera trap images is impacted by training data quality and quantity

Peggy A. Bevan[1,2*], Omiros Pantazis[1,3*], Holly Pringle[1], Guilherme Braga Ferreira[1,4], Daniel J. Ingram[5], Emily Madsen[1,6], Liam Thomas[1], Dol Raj Thanet[7], Thakur Silwal[7], Santosh Rayamajhi[7], Gabriel Brostow[3], Oisin Mac Aodha[8], Kate E. Jones[1]

*Equally contributing authors

1. Centre for Biodiversity and Environment Research, Department of Genetics, Evolution and Environment, University College London, London, UK
2. Institute of Zoology, Zoological Society of London, London, UK
3. Department of Computer Science, University College London, London, UK
4. Instituto Biotrópicos, Diamantina, Brazil
5. Durrell Institute of Conservation and Ecology (DICE), School of Natural Sciences, University of Kent, Canterbury, UK
6. Wildlife Conservation Research Unit, Department of Biology, University of Oxford, Oxford, UK
7. Institute of Forestry, Tribhuvan University, Kathmandu, Nepal
8. School of Informatics, University of Edinburgh, Edinburgh, UK

Corresponding Author: Peggy Bevan, peggy.bevan.17@ucl.ac.uk

**Word Count:**

Abstract: 269

Main Text: Intro = 945, Methods = 2009, Results = 966, Discussion = 1194, Total: **5114**

Acknowledgements: 182

References = 2225

Tables & Figure Legends: 100 + 133 + 107 + 110, Total: 450

**Number of Tables & Figures**: 4

**Abstract**

Large image collections generated from camera traps offer valuable insights into species richness, occupancy, and activity patterns, significantly aiding biodiversity monitoring. However, the manual processing of these datasets is time-consuming, hindering analytical processes. To address this, deep neural networks have been widely adopted to automate image labelling, but the impact of classification error on key ecological metrics remains unclear. Here, we analyse data from camera trap collections in an African savannah (82,300 labelled images, 47 species) and an Asian sub-tropical dry forest (40,308 labelled images, 29 species) to compare ecological metrics derived from expert-generated species identifications with those generated by deep learning classification models. We specifically assess the impact of deep learning model architecture, the proportion of label noise in the training data, and the size of the training dataset on three key ecological metrics: species richness, occupancy, and activity patterns. We found that predictions of species richness derived from deep neural networks closely match those calculated from expert labels and remained resilient to up to 10% noise in the training dataset (mis-labelled images) and a 50% reduction in the training dataset size. We found that our choice of deep learning model architecture (ResNet vs ConvNext-T) or depth (ResNet18, 50, 101) did not impact predicted ecological metrics. In contrast, species-specific metrics were more sensitive; less common and visually similar species were disproportionately affected by a reduction in deep neural network accuracy, with consequences for occupancy and diel activity pattern estimates. To ensure the reliability of their findings, practitioners should prioritize creating large, clean training sets and account for class imbalance across species over exploring numerous deep learning model architectures.

## Introduction

To track progress towards conservation targets (CBD, 2022), there is a pressing need for the intensification of biodiversity monitoring efforts at a global scale (Affinito et al., 2025; Gonzalez et al., 2023; Scharlemann et al., 2020). Steps towards standardised, large-scale monitoring are being introduced through the use of passive monitoring sensors that can scale up data collection efforts (Pimm et al., 2015; Pringle et al., 2025; Stephenson, 2020). Passive biodiversity monitoring sensors like camera traps (e.g. Lee et al., 2024), acoustic monitoring devices (e.g. Sethi et al., 2024), and satellite imagery (e.g. Wu et al., 2023) have allowed researchers to expand their ecological surveys both temporally and spatially with lower field costs and minimal environmental disturbance (Browning et al., 2017). Furthermore, the use of such devices provides standardised and reproducible survey methods and data formats that facilitate collaboration across projects and a network of sensors, slowly forming a global monitoring system (Blount et al., 2021; Kays et al., 2020; Steenweg et al., 2017; Wall et al., 2014).

Specifically, the utilisation of camera traps has been beneficial for monitoring medium to large bodied terrestrial animals, primarily mammals (Burton et al., 2015; Fisher, 2023). These autonomous, motion-activated cameras can be used to collect a variety of ecological metrics such as occupancy (MacKenzie et al., 2002), abundance (Karanth & Nichols, 1998; Rowcliffe et al., 2008), and activity levels (Rowcliffe et al., 2014), which can be used to investigate complex interactions between wildlife, the environment and human activity (Barcelos et al., 2022; Lee et al., 2024; Parsons et al., 2022) and to monitor the success of conservation interventions (Ferreira et al., 2020; Ferreira et al., 2023; Tobler et al., 2015). However, a major limitation of camera trap surveys is the data processing bottleneck, as millions of images need labelling (Duggan et al., 2021; Thomson et al., 2018). This bottleneck causes substantial delays in translating camera trap images into information that can be used in conservation efforts (Merkle et al., 2019).

Machine learning (ML) can increase the efficiency of camera trap analysis and speed up the extraction of ecological information. Deep neural networks have been applied to camera trap and other image data to tackle wildlife monitoring tasks such as locating and identifying species (Miao et al., 2021; Tabak et al., 2019; Willi et al., 2018), counting individuals (Norouzzadeh et al., 2021), classifying behaviour (Kholiavchenko et al., 2024; Norouzzadeh et al., 2018) and estimating occupancy (Whytock et al., 2021). The creation of a general animal detector (MegaDetector) (Beery et al., 2019) has significantly improved efficiency by filtering out empty images, drastically reducing the number of images to be labelled (Fennell et al., 2022; Penn et al., 2024). Other works have exploited the context that accompanies camera trap images to improve species classification directly (Beery et al., 2020) or to learn representations in an unsupervised manner to reduce the number of labels required for training (Pantazis et al., 2021). Despite the growing number of ecological projects that utilise deep neural networks for image classification, evaluation is typically performed through metrics such as the total, or species level, classification accuracy. However, it remains untested whether classification accuracy of a deep learning model is correlated with accuracy of downstream ecological metrics that the detection data are used for.

There is some evidence that ecological information obtained using deep neural networks is comparable to those generated by expert-labelled data. For example, Whytock et al. (2021) found that ML-generated species labels produced similar estimates of species richness, estimated occupancy and activity patterns as expert-labelled data. However, the dataset focused on four species from central Africa, so the spatial and taxonomic generality of the findings are unclear. Practitioners developing deep learning models for species classification must make a series of decisions with respect to the classification model and the training dataset, often constrained by limited compute resources, time to annotate images, or ability to review existing labels. Therefore, even though it has been shown that models with deeper architectures with a higher number of parameters (He et al., 2016), large training datasets (Deng et al., 2009; Lin et al., 2014), and a low proportion of noise in the training dataset (Rolnick et al., 2017; Sukhbaatar et al., 2015) benefit the classification accuracy of a deep neural network, it is unclear what the impact of such factors have on downstream ecological metrics.

Here, we analyse camera trap data from two ecosystems, African Savannah (Maasai Mara, Kenya) and Asian sub-tropical dry forest (Terai region, Nepal) to compare ecological metrics derived from expert-generated image labels with those generated by a trained deep neural network. We specifically assess the impact of neural network model architecture and depth, training dataset size, and proportion of noise in the training dataset on producing three key ecological metrics: species richness, occupancy and activity patterns. It is expected that as these manipulations reduce the classification accuracy, the resulting ecological metrics will deviate further from those produced from expert-labelled data. We expect there may be some robustness in species richness and occupancy, as these only require one positive detection per survey occasion to contribute to the metric. However, activity patterns may be impacted more strongly by a reduction in model accuracy due to the high temporal resolution in the underlying detections. We also explore the relationship between conventional ML evaluation metrics (Top-1 Accuracy, Precision, Recall and F1 score) and accuracy of ecological metrics. Given the shortage of time and resources typically associated with a conservation project, it is not realistic for practitioners to optimise for every parameter. Through our analysis, we aim to shed light on the corresponding impact such factors have on the accuracy of downstream ecological analysis to aid practitioners in their decision making.

## Methods

### Camera trap data

We collected camera trap data from two ecosystems, African Savannah (Maasai Mara, Kenya) and sup-tropical dry forest (Terai region, Nepal). The field sites were part of the Biome Health Project, a field-based study system investigating the impact of human pressures and conservation interventions on biodiversity (https://www.biomehealthproject.com/). Each field site covers a gradient of anthropogenic pressure, but the type of pressure varies between ecosystems (Ingram et al., 2021). Both survey sites were set up using the same survey design. At each field site, un-baited Browning Dark Ops 2017 cameras were deployed evenly across a grid of 2 $km^2$ cells. A single camera was placed as close as possible to the centroid of each survey grid cell and were not biased towards trails or roads. Cameras were attached to a tree or a post at a height of ca. 50cm, were operational 24 h/day with a 1s delay between sequential triggers.

### Maasai Mara Camera Traps (MMCT)

Data was collected from 176 camera traps deployed in four protected areas in the Maasai Mara, south-western Kenya: the Mara Triangle (MT), Mara North Conservancy (MN), Olare-Motorogi Conservancy (OMC), and the Naboisho Conservancy (NB), which each have different restrictions on livestock grazing and human activities. The data were collected throughout October and November 2018 and contains images from 47 species, or groups of species. The data used in this analysis were collected between 5th October and 29th November 2018 and have previously been published in (Connolly et al., 2025). Human infrastructure data were obtained from Klaassen and Broekhuis (2018), and used to calculate the shortest distance from each camera trap station to human development, including settlements, bomas, towns, dams and agriculture (Connolly et al., 2025).

**Bardia Camera Traps (BCT)**

148 cameras were deployed across three contiguous areas under different land management regimes in south-western Nepal: Bardia National Park (NP), the Buffer Zone (BZ) and outside the Buffer Zone (OBZ). These three areas vary in the level of restriction of human activities and development. The survey area therefore covers a gradient of pressure in the form of increasing habitat fragmentation, human density and agricultural activities. The data used in this analysis were collected between 13th February and 16th April 2019 and have previously been published in Ferreira et al. (2023).

**Data Labelling**

We labelled both camera trap datasets by identifying species in each image using the Visual Object Tagging Tool (VOTT) (Microsoft, 2023). Before labelling, images were systematically sampled by using a set time interval of five minutes for MMCT and one minute for BCT, to avoid labelling the same event multiple times (Connolly et al., 2025; Ferreira et al., 2023). The time intervals differed between datasets due to environmental differences between the two ecosystems; the Masai Mara is dominated by large herds of herbivores which consistently triggered the camera, creating an image dataset an order magnitude larger than the BCT dataset for the same survey effort. During labelling, bounding boxes were drawn around each animal, vehicle, or human present in a photo. Images that were hard to identify were shared amongst the labelling team for a second opinion. After labelling, tagging accuracy for each species was checked by randomly sampling 10% of images per species, and any species with poor sampling accuracy in the sample (>3% error rate) were entirely relabelled. This ensured that both manually labelled datasets are highly accurate and contained minimal errors. To account for the fact that some species were under-represented within our collected data, relevant and visually similar species were grouped together where necessary, resulting in a list of labels that consists of either species or species groups (Table S1; Table S2).

For deep neural network training, the labelled images from each dataset were split into subsets for model training, model validation and testing model performance. The dataset was split temporally, not accounting for class, to ensure each subset had even spatial coverage across the survey area. Due to the shorter collection times of these datasets (2 months for MMCT; 1 month for BCT), seasonality did not need to be considered when creating these subsets. The MMCT dataset was split into 53,102 images for model training, 5,879 images for validation, and 23,319 images for testing model performance. The BCT dataset was split into 28,210 training, 3,119 validation, and 8,979 test images. For the final ecological analyses, we performed a series of filtering steps by first applying a minimum threshold of 20 expert labels per species in the test dataset, as this allowed a reasonable classification accuracy to be quantified. We also removed domestic species, birds and any species groups that contained combinations of visually similar species (e.g. small cat category in BCT contained domestic cat and jungle cat). This resulted in 20 mammal species from the MMCT dataset and 8 mammal species from the BCT dataset (Table S3).

## Ecological Analysis

### Species Richness

Species richness was measured as the count of wild species observed at each camera trap location (a single camera trap deployment) over the entire survey period, using species detections from either the expert-generated labels or labels predicted by a deep neural network.

### Occupancy

We adopted a multi-species occupancy framework to estimate occupancy while accounting for imperfect detection (Dorazio et al., 2006). Given the differences between the two study areas, we implemented slightly different occupancy models for the MMCT and BCT datasets (see Supplementary Text 1 and 2 for model specifications). To quantify the impact of deep neural network-based image classification on ecological responses, we investigated the effect of variables that had a strong influence on occupancy according to the model results: shortest distance to human infrastructure in the MMCT dataset and management regime in BCT dataset (following Ferreira et al., 2023). We ran both occupancy models on the species detections from either expert-generated labels or labels predicted by a deep neural network and extracted the species-specific model coefficients with 95% credible intervals from the posterior distributions.

The model coefficients represent the effect of a variable on occupancy, *i.e.* the response of zebra occupancy to distance from human infrastructure. For the MMCT occupancy model, we extracted the raw coefficients of distance to human infrastructure as the response, as this is a continuous metric. For the BCT occupancy model, 'management regime' is a categorical variable, so we calculated the difference in occupancy probability between National Park (NP) and Outside Buffer Zone (OBZ) as the response to changing management regime.

**Activity Patterns**

We estimated the diel activity pattern of each species by fitting a circular kernel density function using the activity R package (version 1.3.4) (Rowcliffe, 2023). The kernel density function calculates the probability a species is active at each moment over a 24-hour period. To avoid large biases in estimates, only species that had ≥ 20 detections were included in the analysis (Rowcliffe et al., 2014). For each species, we calculated activity patterns from detections from expert-labelled data and the deep neural network, and compared them by calculating the bootstrapped overlap coefficient of the two activity patterns using the overlap R package (version 0.3.4) (Meredith & Ridout, 2023). The resulting overlap coefficient ranges from 0 to 1, where a value of 1 means perfect overlap between the two activity patterns. In this case, a higher overlap value indicates high ecological accuracy of the deep neural network-generated labels when compared to the expert labelled data.

**Deep Neural Network Experiments**

To investigate the impact of deep neural networks on downstream ecological metrics, three experiments were run, each of which manipulated a different aspect of the training pipeline. We varied the underlying model architecture, the size of the training set, and the proportion of noise (incorrect labels) within the dataset. Except for the model architecture experiment, the baseline model utilised for each experiment is a ResNet50 CNN (He et al., 2016). This model is commonly used as a baseline in machine learning experiments. All experimental deep neural network models were sourced from the PyTorch library (Paszke et al., 2019). Across all experiments we utilised transfer learning (Yosinski et al., 2014), where neural network models were initialised from weights obtained via pre-training on ImageNet (Deng et al., 2009). This approach has demonstrated benefits for biodiversity monitoring by improving model accuracy (Willi et al., 2018). The model training was conducted on crops of each animal image (determined by the bounding box drawn in the labelling process) given that this approach has shown to benefit model accuracy (Beery et al., 2018; Gadot et al., 2024; Norman et al., 2023). We trained the classifier on all classes from the training set (Table S1; Table S2), but we only make predictions for the subset of classes that belong to the list of species on which the ecological analysis is conducted (Table S3). For evaluation of trained deep neural networks, we predicted labels on a held-out test set and applied a 70% confidence threshold across all experiments, as filtering out uncertain labels has been shown to improve the robustness of the resulting ecological analysis (Whytock et al., 2021). The deep neural network predictions are then treated in the same way as expert-labelled datasets and used to calculate the ecological metrics described above as well as conventional ML evaluation metrics.

**Impact of Deep Learning Classification Model**

To examine the impact that model architecture and depth has on downstream ecological metrics, four model types were compared. These were three ResNet (He et al., 2016) models with varying depth, ResNet18, 50 and 101. We chose to use the ResNet architecture because of its widespread use in computer vision research on camera trap data (Beery et al., 2018; Willi et al., 2018). The inclusion of three ResNet models at multiple depths allows us to explore the impact of using deeper models. In addition, a ConvNeXt-T model was used. This CNN uses a modified version of a ResNet architecture, that takes inspiration from state-of-the-art vision transformer (ViT) models to achieve a higher performance with almost half the number of parameters as ResNet101 (Liu et al., 2022). The ConvNeXt-T model has similar performance to many ViT variants with similar parameter counts (Pucci et al., 2023; Vishniakov et al., 2024).

**Impact of Training Set Label Noise**

To investigate the impact of label noise (incorrect labels) on downstream ecological metrics, we created six versions of each training set (MMCT and BCT) with varying levels of label error, from 1% to 50% of examples mis-labelled. For a realistic simulation of label errors within each species, the iNaturalist citizen science platform (iNaturalist, 2023) was used to retrieve the three species that were most commonly misidentified as the original species, that also exist in our data (Table S4). In the case that three species were not available from iNaturalist, another species from the dataset was randomly selected.

**Impact of Training Set Size**

To investigate the impact of training set size on downstream ecological metrics, 7 versions of the training set were created for each dataset (MMCT and BCT) where the number of labels for each species was varied, from 100% of the original training set to 1% of the original training set.

### Correlation Between Machine Learning Evaluation and Ecological Metrics

To describe the relationship between deep neural network accuracy and ecological metric accuracy, we measured the correlation between classification error and ecological accuracy. To quantify 'ecological accuracy', we took the absolute difference between the species-level occupancy coefficients measured from expert-generated labels and deep neural network-generated labels. For activity patterns, 1 minus the species-level overlap value was used, so that for both metrics, a 0 value equates to perfect prediction i.e. no deviation between expert-labelled data and classifier prediction. To quantify classification error, we use four metrics commonly utilised to evaluate deep neural network classification performance: Top-1 Accuracy (proportion of correct classifications among all images), Precision (proportion of true positives among all positive predictions), Recall (proportion of true positive predictions out of all actual positives, including false negatives), and F1 Score (harmonic mean of Precision and Recall). To test correlation with ecological metrics, we use the error rates of these metrics i.e. Top-1 Error, Precision Error, Recall Error, and F1 Error, which are calculated as 1 − metric.

## Results

### Species Richness

We found that species richness predicted from DL-generated labels was robust to different model architectures and model depths but was impacted by high levels of noise in the training data and reduction in training set size, particularly in the BCT dataset which is relatively smaller (Fig. 1a, d). Each of the four DL models accurately estimated the number of species present at each location compared to expert-labelled data. Although the error was slightly higher in the BCT dataset than the MMCT dataset, this did not appear to be driven by any particular DL model. In contrast, when there is more than 10% label noise in the dataset, species richness is underestimated (Fig. 1b, e). Reduced training set size also had an impact on underestimating species richness, but to different degrees between the datasets. For the MMCT dataset, an impact of reduced training set on predicting species richness is clear at 10% of the original size (Fig. 1c). However, for the BCT dataset, this impact is clear from 25% and below (Fig. 1f).

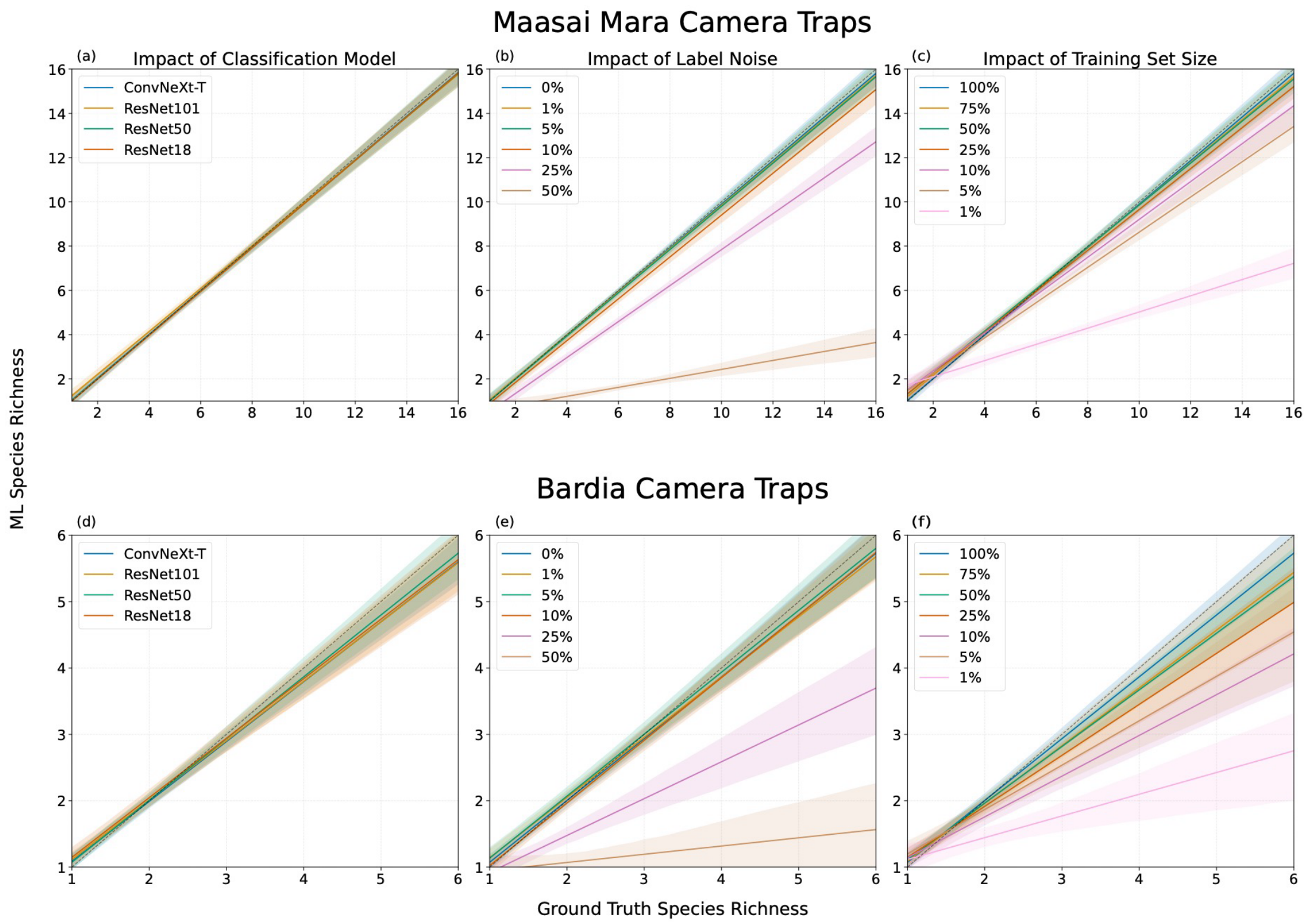

Fig. 1. Observed species richness calculated with a variety of deep neural network architectures and training settings across two datasets. The Y-axis corresponds to the observed species richness per camera trap location using labels predicted by a deep neural network and the X-axis corresponds to species richness calculated using labels provided by experts. The lines in each plot correspond to a linear fit of the calculated richness across camera trap locations and a diagonal line that goes through the origin corresponds to the perfect match between the two axes. Shaded areas show 95% confidence intervals for the linear regression model.

## Occupancy Modelling

The results of occupancy modelling showed that species' labels predicted from deep neural networks can recover some species-specific responses to environmental covariates, even with high levels of manipulation to the training set (Fig. 2a). However, responses of less common and visually similar species were not predicted consistently. In the MMCT occupancy model derived from expert-generated labels, four species had posterior estimates indicating higher occupancy at sites further from human infrastructure, 13 species showed little evidence of an effect, and 3 species has estimates suggesting lower occupancy (Fig 2a). Posterior distributions for elephant (n = 549 training images) and zebra (n = 4424 training images) consistently indicated positive effects across all DL models, even with up to 10% label noise and 10% of the original training set size. These effects were not evident for eland (n = 407 training images) and inconsistent for topi (n = 1440 training images). Some species had coefficients with 95% credible intervals overlapping zero in occupancy models from expert-generated labels (e.g. hippopotamus) yet showed strong posterior evidence for positive effects in many of the DL-predicted datasets. This pattern was more common for species with negative coefficients, such as vervet monkey (n = 382 training images) and Grant's gazelle (n = 543 training images). Grant's gazelle is visually similar to Thomson's gazelle (n = 5580 training images), which consistently showed posterior estimates indicating a negative effect across multiple manipulations, suggesting detections of these two species were conflated by our algorithms.

Occupancy results from the BCT dataset were slightly more consistent overall. Posterior estimates from the expert-generated labels indicated three species had higher occupancy in NP (chital, grey langur and sambar), four species had little evidence of a difference between NP and OBZ, and one species had higher occupancy in OBZ (nilgai) (Fig 2b). The responses of chital (n = 8715 training images) and grey langur (n = 391 training images) were consistently recovered in occupancy models from DL-generated labels, even with high levels of label noise and up to 5% of the original training set (Fig. 2b). However, the response of sambar (n = 265 training images) was less consistently recovered, and nilgai (n = 185 training images), one of the least common species in the dataset, was never correctly recovered. Notably, the posterior support of a positive response in sambar was missed by our baseline model (ResNet50) but was occasionally recovered by our deep neural networks under heavy manipulation. Further exploration found that these confusing responses were likely to be a false signal: sambar classification accuracy declined with increasing manipulations and the range of species misclassified as sambar increased, suggesting false positives were driving spurious occupancy patterns. Similar effects were seen for barking deer and wild boar.

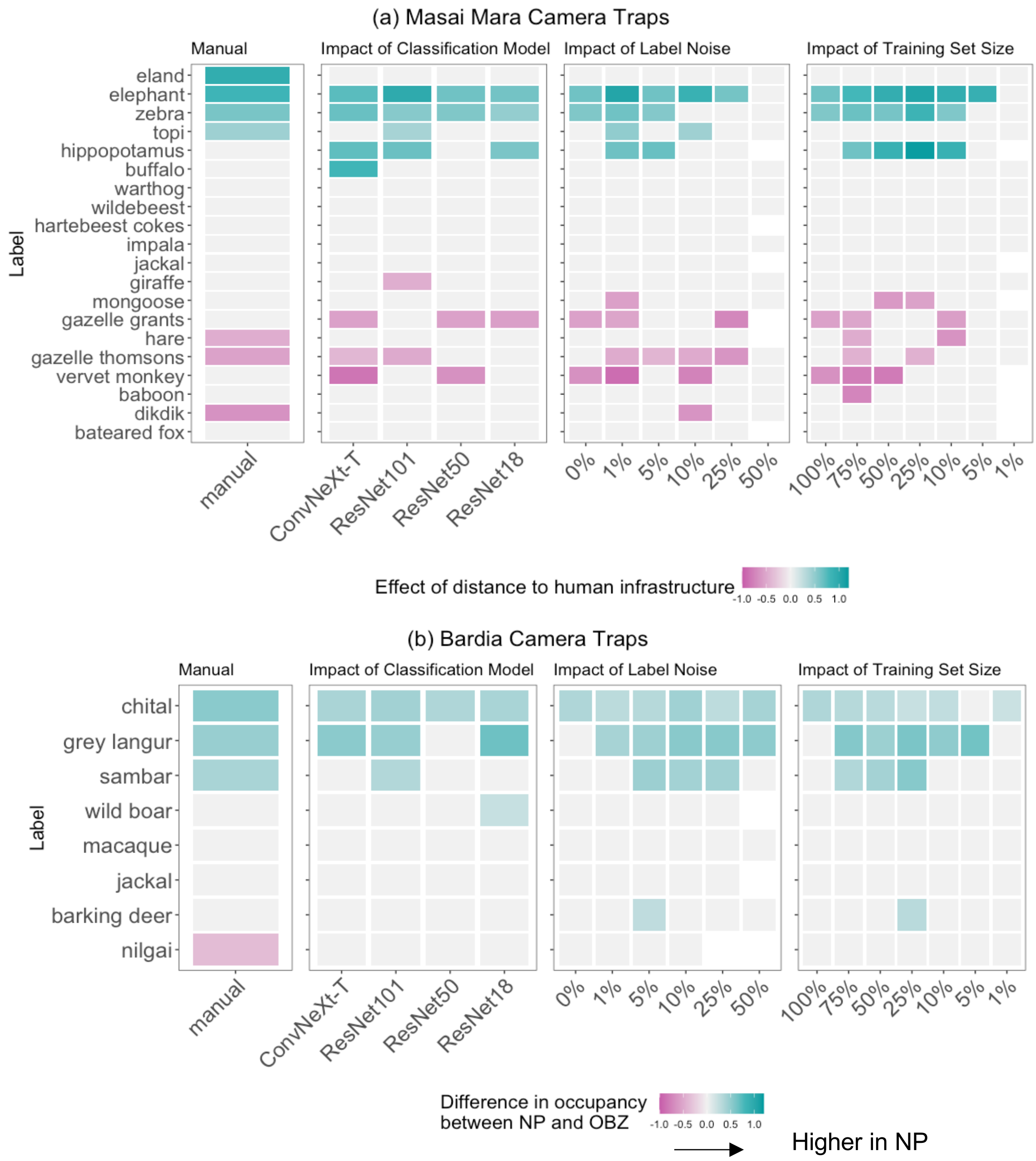

Fig. 2. Responses of occupancy to environmental pressures predicted from expert-generated (manual) labels and deep neural network-generated labels with various manipulations in the training pipeline. Plots show results of multi-species occupancy model predicting the species-specific occupancy response of mammals to a) distance to human infrastructure in the Masai Mara, Kenya, where a positive effect means higher occupancy further away from infrastructure, and b) to management regime in Bardia, Nepal, where a positive effect means occupancy is higher in the National Park (NP) than in unprotected land (OBZ). Each tile shows the direction of the effect indicated by posterior estimates, and grey tiles show cases where the 95% credible interval overlapped zero, suggesting little evidence of an effect. The X-axis represents a range of manipulations to model architecture and training set.

**Activity Patterns**

The predictions from all four DL models produced a similar range of activity pattern overlap with the expert-labelled data, with most overlap coefficients for the species ranging between 0.8 and 1 (Fig. 3a, d). Even though ConvNext-T performs well for most of the species, on the BCT dataset it detected fewer species when compared to the ResNet models; one species was dropped from analysis due to lack of detections (Fig. 3d). Average overlap in the MMCT dataset is consistent with changing model architecture, but there is a clear trend of reducing overlap with model architecture in the BCT dataset, with the ResNet18 performing worst out of the four models (Fig. 3d). Accuracy of activity patterns were robust to a certain level of noise, with reductions in number of species and overall accuracy beyond 10% noise in both datasets (Fig. 3b, e). The activity patterns were also robust to a 50% drop in training set size (Fig. 3c, f). Displaying the predicted activity patterns of a subset of species showed that model manipulations mainly caused an overestimation of diurnal predictions compared to expert-labelled data (Fig S1; Fig S2).

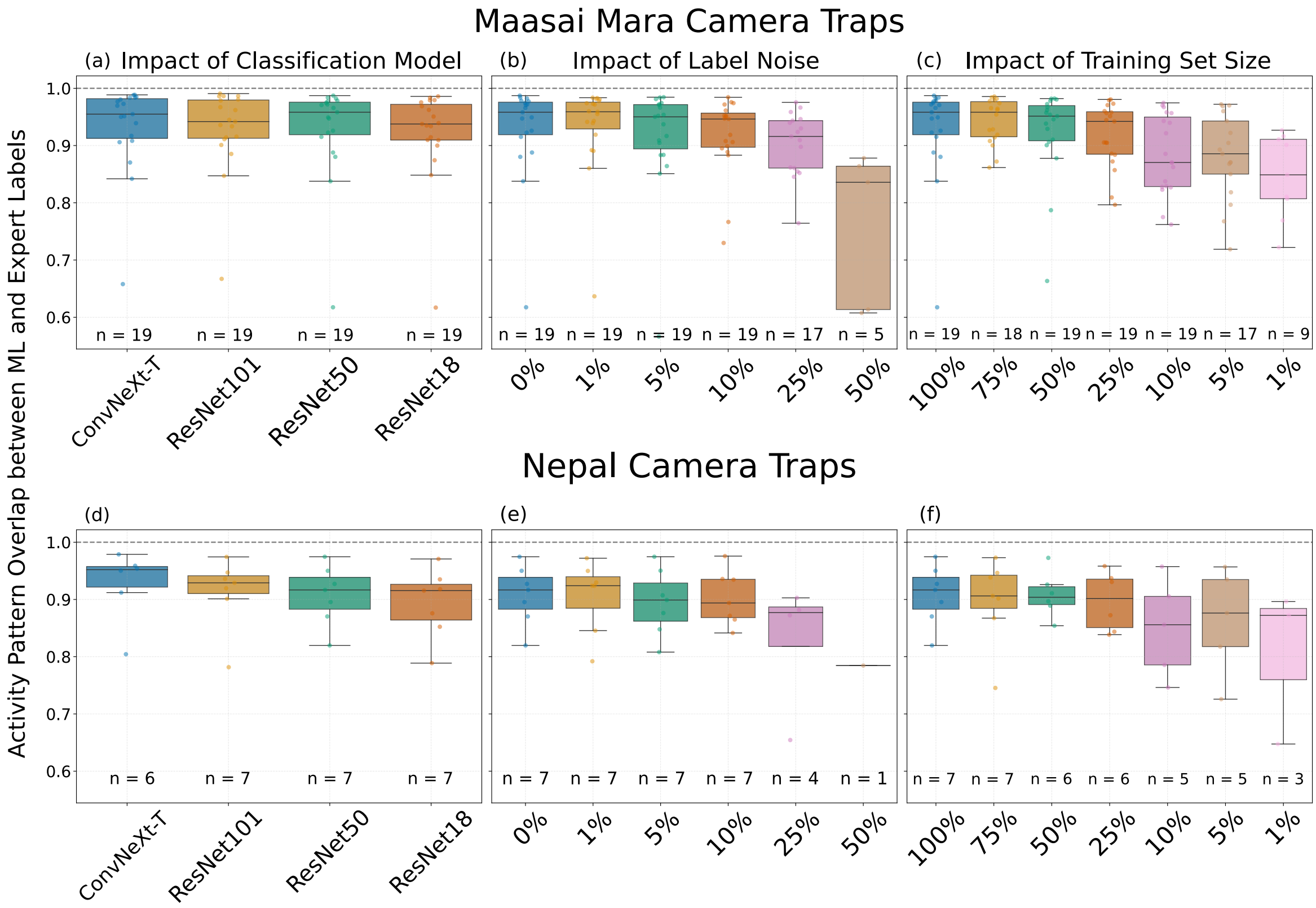

Fig. 3. The overlap coefficient of species activity patterns produced from expert - generated labels compared to deep neural network-generated labels with variety of manipulations in the training pipeline. The box plots describe the mean, upper and lower quartile of the overlap coefficients calculated for each species in each dataset. The Y-axis corresponds to the activity overlap coefficient, where a value of 1 represents perfect agreement between the activity patterns calculated from expert-generated labels and DL-generated labels. In cases where the deep neural network did not predict ≥ 20 detections of a species, this overlap calculation was dropped from the aggregate, represented in the n under each boxplot.

## Correlation Between Machine Learning Evaluation and Ecological Metrics

Classification accuracy of each deep neural network reduced with increasing manipulations, as expected (Table S6). However, we found that the accuracy of ecological metrics derived from deep neural network-generated labels is not always directly correlated with performance of the deep neural network. In most cases, the error in predicting ecological metrics is quite low, for example when looking at the impact of model, the majority of ecological accuracy values are below 0.4 (0 = perfect accuracy) (Fig. 4a, d). Overall, the accuracy of activity pattern predictions is more strongly correlated with deep neural network performance than the accuracy of occupancy (Fig. 4). Classification model and reducing training set size had a weak positive correlation with occupancy accuracy, but when looking at the impact of increasing noise in the dataset on occupancy predictions, there is no correlation with model performance (Fig. 4b, e).

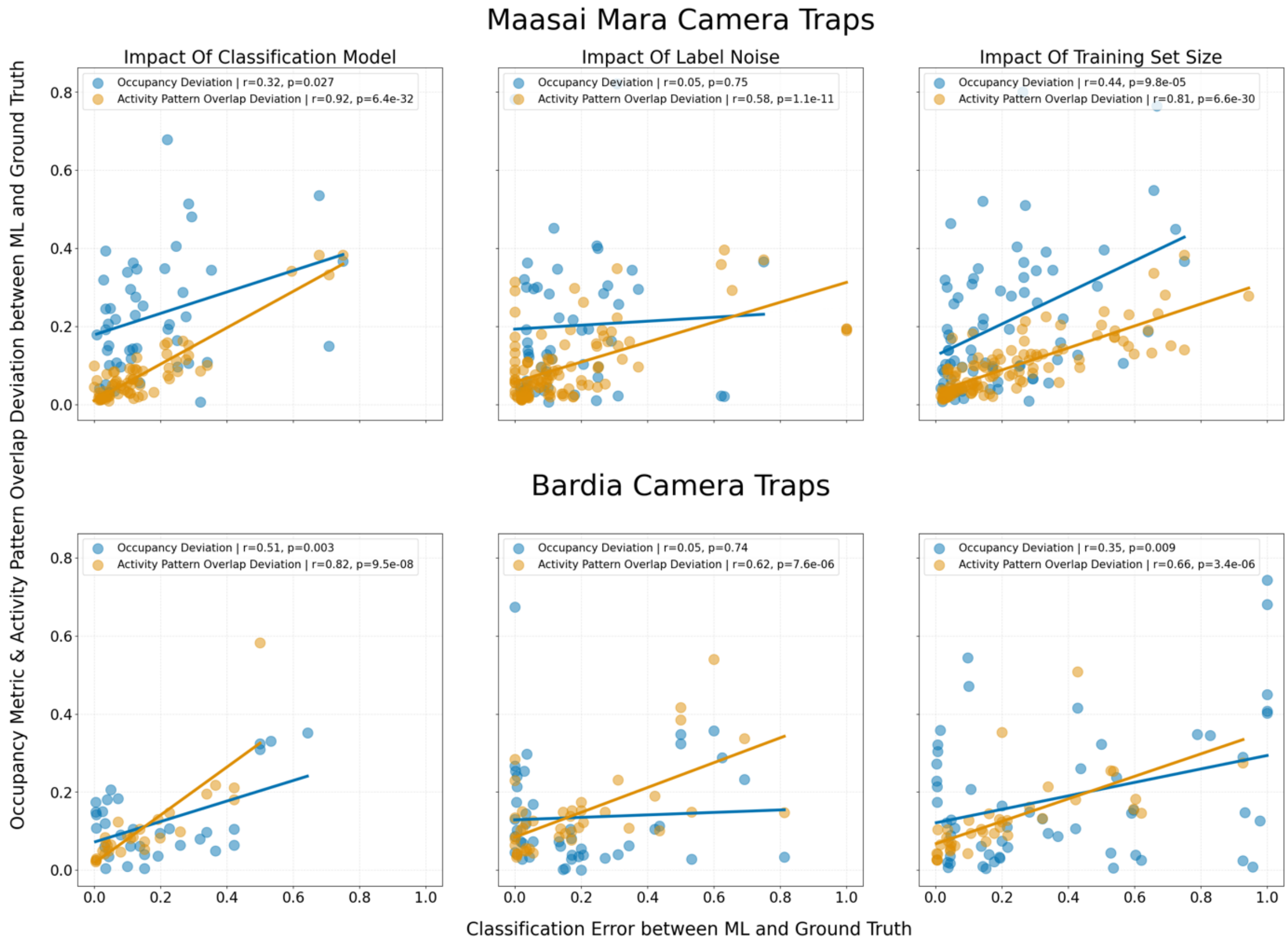

Fig. 4. Correlation between ecological metric accuracy and deep neural network classification accuracy. Ecological metric accuracy is calculated as the absolute difference between metrics calculated from neural network-generated labels and expert-generated labels. The Y-axis corresponds to the absolute difference between occupancy coefficient estimate (blue lines) or activity pattern overlap (orange lines). The X-axis shows the classification error of each neural network. For each experiment, the individual manipulations are pooled together to show the overall trend, for example the different DL models used in the Impact of Classification Model experiment are not differentiated. The results cover 20 and 8 species from MMCT dataset and BCT dataset respectively.

## Discussion

Our study shows that deep neural network species classifiers can serve to estimate ecological metrics with reasonable accuracy, particularly for community-level measures like species richness. However, estimates were less reliable for rarer or visually similar species. The utilisation of two camera trap datasets from two very different biomes, African savannah (MMCT) and Asian sub-tropical dry forest (BCT), each with a high diversity of medium and large mammals increases the generality of our findings. Despite differences between ecosystems, background, dataset size, and species composition, our findings were broadly consistent across datasets. Using two datasets with standardized survey design and rich metadata allowed us to control for survey effort and site-specific context, enabling robust analyses of occupancy and activity patterns that would have been difficult with benchmark datasets such as iWildCam (Beery et al., 2021), illustrating the value in using real, location-specific data when testing how machine learning can support biodiversity monitoring.

### Deep Neural Network Experiments

We found that that the choice of image classification model architecture or model depth has very little impact on the resulting ecological findings. Even though the utilisation of deeper (He et al., 2016) or better performing (Liu et al., 2022) model architectures results in increased classification performance, we observe that CNN architectures that may be considered lower performing or outdated, such as a ResNet50, can still produce accurate predictions of species richness or occupancy for some species. Indeed, more advanced neural networks architectures than ResNet or ConvNext-T are now available. However, the purpose of this study was to assess the relative influence of model depth and architecture type on downstream ecological analyses. This insight can help practitioners to allocate available resources more efficiently, as there may not be a need for large models with expensive computational needs.

Our experiments show that most ecological metrics calculated from DL-predicted labels maintained a high similarity with expert-labelled data, even with up to 10% noise in the training set - a pattern that is common for deep neural network approaches (Drory et al., 2018; Rolnick et al., 2017). Above levels of 10% noise, reduced prediction confidence caused by noisy training labels means that more labels are dropped when a 70% confidence threshold is applied, resulting in apparent non-detection of rarer species. This created a bias in the results presented, as the measure of metric accuracy does not account for missing species. The loss of rarer or more cryptic species from a dataset represents a wider problem of class imbalance that is regularly seen in camera trap datasets and can disproportionately affect detection of less common species (Schneider et al., 2020). Additionally, CNNs are more resilient to noise that is uniformly spread across the dataset, compared to noise that is concentrated (Drory et al., 2018). A potential hindrance to our experimental design is the choice to mis-label images uniformly across species, when real labelling error would likely be focussed on a particular species or part of the dataset (*e.g.* nocturnal images only).

The accuracy of the key ecological metrics generally held up against reductions in the training set size (*e.g.* up to 50% of the available data dropped) across both datasets. We note that we applied a temporal data split to maximise applicability to this dataset, when a spatial split might be more conventional (Norman et al., 2023). In that scenario, reducing the training set size could more strongly impact generalization to new sites.

Past research has succeeded in building accurate classification models when training with millions of images, a scenario that might not be possible in most wildlife monitoring projects given the time-consuming and demanding nature of the manual image annotation step (Norouzzadeh et al., 2018). Thus, knowing where to stop within the labelling phase is a crucial decision given the time-consuming nature of camera trap image annotation. The emergence of efficient methods within the biodiversity monitoring domain such as active learning (Norouzzadeh et al., 2021), self-supervised learning (Pantazis et al., 2021), or large vision-language models (Pantazis et al., 2022) claim to reduce the need for large labelled sets. Whilst it is clear from our results that large training sets will always improve classifier accuracy, using these cutting-edge methods will further reduce the need for image labelling.

Even though our results suggest a small amount of label noise or reduced training set size is acceptable when predicting community-level metrics such as species richness, we observed a disproportionate impact on less common species for species-specific metrics. Our occupancy model results showed that classification error led to spurious responses of certain species to human infrastructure in Kenya or protected area management in Nepal. All analyses showed that certain species were dropped completely due to a lack of detections. This demonstrates how misclassifications caused by lower performing deep neural networks could obscure ecological patterns and misinform conservation decisions if interpreted as reliable signals. Our findings highlight the need to consider class imbalance when reporting on ecological analysis from DL-generated labels. Addressing such problems could involve species-specific confidence thresholds, statistical methods that account for false positives and false negatives (Katsis et al., 2025; Royle & Link, 2006), or developing nested classification models that focus on a subsets of animals (Mulero-Pázmány et al., 2025).

## Correlation between conventional Neural Network Evaluation and Ecological Metrics

We observe that the accuracy of deep learning-based ecological analysis does not always correlate strongly with conventional ML evaluation metrics, but this varied with model manipulation and ecological metric. For example, adding noise to training labels degrades ML classification performance but there is no analogous impact on estimating species' occupancy responses to anthropogenic pressure, although a weak correlation appeared when reducing training set size. Classification error had a greater impact on activity pattern accuracy. This may be due to the resolution of detections needed for each analysis. For species richness and occupancy, detection frequency at a CT site within the detection window does not affect the metric, allowing for a degree of classification error. Predicting activity patterns requires several detections over the 24hr cycle, and any loss, *e.g.* less detections at night, strongly affects interpretation. Dataset characteristics will also impact interpretation: the MMCT dataset is dominated by grazers, many of which are visually similar with overlapping ecology. This could not be said for the BCT dataset, where species show a diversity of responses to changing management regime (Ferreira et al., 2023), making accuracy of responses more sensitive to label accuracy. It is important for practitioners to understand this impact when choosing analysis tools and to remain transparent in the DL methods used.

## Conclusion

This study presents an end-to-end evaluation of deep learning models trained under different settings based on metrics relevant to downstream ecological tasks. Our results provide clarity on the robustness of such models against a variety of typical design decisions related to DL model training and highlight areas where caution is warranted, for example interpreting species-specific responses, particularly for rare or visually similar species. Ultimately, our findings aim to empower practitioners with limited access to high computing power or specialist knowledge to build effective tools for conservation. Future research in this field should focus on enhancing accessibility, ensuring that deep learning tools can be widely adopted and applied by the global conservation community.

**Acknowledgements:** This research was funded by WWF-UK as part of the Biome Health Project. Peggy Bevan acknowledges support by the Natural Environment Research Council (NERC), United Kingdom (Grant Ref.: NE/S007229/1). We thank Miranda Jones, Sarah Carroll, Georgia Cronshaw, Stratton Hatfield, Alex Rabeau, Taras Bains and Liam Patullo for their help with image tagging. Thank you to Naresh Khanal, Prabin Poudel, Thomas Luypaert, Tilak BK and several students from Tribhuvan University who helped with data collection. Research permits were granted by the Nepali Ministry of Forests and Environment and the Kenyan National Commission for Science, Technology and Innovation (NACOSTI) (Ref. no: NACOSTI/P/18/61494/23703). Research permit was also granted by the Kenya Wildlife Services (Ref.: KWS/BRM/5001). We acknowledge WWF Kenya for logistical support and Olare Motorogi, Naboisho, Mara North and the Mara Triangle conservancies and their teams in Kenya for access and field support. Daniel J. Ingram acknowledges support from UK Research and Innovation (Future Leaders Fellowship, Grant Ref.: MR/W006316/1). This article is written in memory of Dr. Ben Collen and Professor Dame Georgina Mace who both sadly passed since the inception of the Biome Health Project.

**Author's Contributions Statement:** *PB, OP, HP and KEJ* conceived the ideas and designed methodology; *DJI, LT & EM* designed the field surveys and data collection; *EM & HP* collected & labelled the MMCT data; *LT, DRT, SR, TS, GBF* and *PB* collected and labelled the BCT data; *OP* conducted all machine learning model training; *PB & HP* conducted ecological analyses; *OMA, GB, KEJ* & *GBF* provided guidance and supervision on data collection, analysis and writing. *PB* and *OP* shared the manuscript writing equally. All authors contributed critically to the drafts and gave final approval for publication.

Our study includes authors from Nepal, who aided in locating a suitable field site, data collection and providing ecological knowledge on the area. We do not include any authors from Kenya as this part of the project did not have any academic partners with capacity to contribute to this manuscript. Team members of the Biome Health Project performed multiple stakeholder meetings involving academics, government officials and NGOs in both countries to disseminate results. The funding for the Biome Health Project has allowed for multiple capacity building activities, including teaching and supervision of MSc students and a PhD studentship at Tribhuvan University, Nepal. We also provided training for rangers in the Maasai Mara area in using camera traps.

**Data Accessibility Statement:** Data have been made available for review through this restricted link: https://zenodo.org/records/13372531. We plan to archive them in Zenodo or LILA BC after publication of this work. Similarly, our code is available for review in the following anonymized repository and will be released publicly in GitHub upon acceptance: https://anonymous.4open.science/r/ml_ecological_metrics-9F54/README.md.